\documentclass[11pt]{article}
\usepackage{eamt23}
\usepackage{times}
\usepackage{url}
\usepackage{latexsym}
\usepackage[small,bf]{caption} 
\usepackage{multirow}
\usepackage{boldline}
\usepackage{adjustbox}
\usepackage[table,xcdraw]{xcolor}
\usepackage{amsthm}
\usepackage{mdframed}
\usepackage{subfig}
\usepackage{enumitem}
\usepackage{pbox}
\usepackage[implicit=false, bookmarks=false]{hyperref}

\newmdtheoremenv[linewidth=2pt, topline=false, bottomline=false, rightline=false,%
leftmargin=0pt, innerleftmargin=0.4em, rightmargin=0pt, innerrightmargin=0pt, innertopmargin=-5pt ,%
innerbottommargin=3pt, splittopskip=\topskip, splitbottomskip=0.3\topskip, %
skipabove=0.6\topsep]%
{example}{Example}%

\title{Tailoring Domain Adaptation for Machine Translation Quality Estimation}

\author{Javad Pourmostafa Roshan Sharami, Dimitar Shterionov, Frédéric Blain,\\ \textbf{Eva Vanmassenhove, Mirella De Sisto, Chris Emmery, Pieter Spronck}\\\\
Department of Cognitive Science and Artificial Intelligence, Tilburg University \\
\texttt{\{j.pourmostafa,d.shterionov,F.L.G.Blain,e.o.j.vanmassenhove,}\\\texttt{M.DeSisto,C.D.Emmery,p.spronck\}@tilburguniversity.edu
}}

\date{}

\begin{document}
\maketitle
\begin{abstract}
While quality estimation~(QE) can play an important role in the translation process, its effectiveness relies on the availability and quality of training data. For QE in particular, high-quality labeled data is often lacking due to the high cost and effort associated with labeling such data. Aside from the \textit{data scarcity} challenge, QE models should also be generalizable; i.e., they should be able to \textit{handle data from different domains}, both generic and specific. To alleviate these two main issues --- data scarcity and domain mismatch --- this paper combines domain adaptation and data augmentation in a robust QE system. Our method first trains a generic QE model and then fine-tunes it on a specific domain while retaining generic knowledge. Our results show a significant improvement for all the language pairs investigated, better cross-lingual inference, and a superior performance in zero-shot learning scenarios as compared to state-of-the-art baselines.
\end{abstract}

\section{Introduction}
\label{sec:intro}
Predicting the quality of machine translation~(MT) output is crucial in translation workflows. Informing translation professionals about the quality of an MT system allows them to quickly assess the overall usefulness of the generated translations and gauge the amount of post-editing that will be required~\cite{tamchyna-2021-deploying,murgolo-etal-2022-quality}. Quality estimation~(QE) is an approach that aims to reduce the human effort required to analyze the quality of an MT system by assessing the quality of its output without the need for reference translations.\looseness=-1

QE can be applied on word-, sentence- or document-levels. The goal of sentence-level QE, which is the focus of our work, is to predict a quality label based on a source sentences and its MT equivalents. This label, (i.e., the quality estimate), can be expressed in various ways such as TER/HTER~\cite{snover-etal-2006-study}, BLEU~\cite{papineni-etal-2002-bleu} or any metric of interest to the user. Training a sentence-level QE system typically requires aligned data of the form: \textit{source sentence} (SRC), \textit{target sentence} (TRG), and \textit{quality gold label} (LBL).
However, most quality labels are by-products of MT and post-editing --- a rather difficult and expensive process --- limiting the size of the available QE data~\cite{rei-etal-2020-comet,Zouhar2023PoorMQ}.\looseness=-1

The WMT QE shared task~\cite{specia-etal-2021-findings,zerva-etal-2022-findings} has been offered a platform to compare different QE systems and to share QE data. 
Despite efforts from initiatives like the QE shared task to publicly release QE datasets, such resources remain scarce across language pairs and, by extension, also have a limited coverage across domains~\cite{2020arXiv201004480F,fomicheva-etal-2022-mlqe}.
This can pose a challenge for all QE models, especially recent ones that utilize large pre-trained language models (LLMs)~\cite{ranasinghe-etal-2020-transquest,zerva-etal-2022-findings}, since fine-tuning pre-trained models with small datasets has been demonstrated to be quite unstable~\cite{2020arXiv200605987Z,rubino-2020-nict}.

Furthermore, QE models trained on specific data do not generalize well to other domains that are outside of the training domain~\cite{kocyigit-etal-2022-better}. \textit{Domain mismatches} lead to significant decreases in the performance of QE models~\cite{c-de-souza-etal-2014-machine,Zouhar2023PoorMQ}. To improve the generalizability of QE models, it is important to establish the right balance between domain-specific and generic training data.
To date, only a few attempts have been made to address this challenge~\cite{de-souza-etal-2014-towards,rubino-2020-nict,lee-2020-two}. Thus, the majority of QE models have difficulty with accurately estimating quality across different domains, whether they are generic or specific~\cite{Zouhar2023PoorMQ}.\looseness=-1

In this work, we propose to tackle both the data scarcity and the domain mismatch challenge that LLM-based QE models face. \textit{We propose a methodology whereby a small amount of domain-specific data is used to boost the overall QE prediction performance.} This approach is inspired by work on domain adaptation~(DA) in the field of MT, where a large generic model is initially trained and then fine-tuned with domain-specific data~\cite{chu-wang-2018-survey,pham-etal-2022-multi}.

To assess the validity of the proposed approach in QE, we conducted experiments using small and large, authentic and synthetic data in bilingual, cross-lingual, and zero-shot settings. We experimented with publicly available language pairs from English (EN) into German (DE), Chinese (ZH), Italian (IT), Czech (CZ), and Japanese (JA) and from Romanian (RO) and Russian (RU) into English (EN). We used the common test sets from the WMT2021 QE shared tasks\footnote{\url{https://www.statmt.org/wmt21/quality-estimation-task.html}}. 

Our experiments show a statistically significant improvement in the performance of QE models.
Our findings also indicate that not only our implementation leads to better multi-/cross-lingual QE models (where multi-/cross-lingual data is provided) but also zero-shot QE (where no data for the evaluated language pairs was provided at training).

The main contributions of our research are:
\begin{itemize}[leftmargin=*,noitemsep]
    \item A QE methodology that employs DA and data augmentation (DAG), along with a novel QE training pipeline that supports this methodology.
    \item An empirical demonstration of the pipeline's effectiveness, which highlights improvements in QE performance, and better cross-lingual inference.
    \item A comparative analysis with state-of-the-art (SOTA) baseline methods that demonstrates the effectiveness of our approach in enhancing zero-shot learning~(ZSL) for the task of QE.
    \item Adaptable QE pipelines that can be tailored and implemented for other language pairs; i.e., high generalizable QE pipelines.
\end{itemize}

To the best of our knowledge, this is the first QE methodology to use DA and DAG. Furthermore, it is easily reusable and adaptable: (i) while we used XLM-R in our experiments, one can easily replace it with any preferred LLM as long as the input-output criteria are met; (ii) we built our tool around Hugging Face~(HF) implementations of LLMs, meaning one can employ a certain generic model and apply it to any QE task by simply fine-tuning it on (newly-collected) QE data.\looseness=-3



\section{Domain adaptation for specialized QE} 
\label{sec:methodology}
In this section, we outline our methodology for training LLM-based QE models for a specific domain with limited available in-domain data. This involves: (i) a set of training steps that we found to be particularly effective, 
and (ii) DAG techniques to improve the QE models' specificity. Additionally, we provide details on two different training modes we implemented (with or without tags).

\subsection{Training steps}
\label{sec:training_steps}
We implement the ``mixed fine-tuning + fine-tuning'' DA technique that proved promising for MT~\cite{chu-etal-2017-empirical}. We tailor this methodology to suit our needs following the steps outlined below. A visualization of the steps involved can be found in~Appendix~\ref{appendix:training_steps}. Our technique involves leveraging both in-domain (ID) and out-of-domain (OOD) QE data (see Section~\ref{sec:data} for details on the datasets). 

\paragraph{Step 1} We train a QE model using OOD data until it converges. We employ the experimental framework described in Section~\ref{sec:QE} in which an LLM is fine-tuned to predict QE labels. The goal of this step is two-fold: (i) leveraging the LLM's cross-lingual reference capabilities and (ii) building a generic QE model. This way we ensure that the model can estimate the quality of a broad range of systems, but with limited accuracy on ID data.
 
\paragraph{Step 2} The model's parameters are fine-tuned using a mix of OOD and ID data. We use different ID data, both authentic and synthetic according to the DAG approaches in Section~\ref{sec:DAG}. The objective here is to ensure the model does not forget generic-domain knowledge acquired during the first step while simultaneously improving its ability to perform QE on the domain-specific data. This mixing step is often referred to as ``oversampling'' in DA literature, where a smaller subset of OOD data is concatenated with ID data to allow the model to assign equal attention to both datasets; it aims to further adapt the model to the specific domain of interest. \looseness=-1

\paragraph{Step 3} We continue to train the QE model on a specific ID dataset until convergence, resulting in a more domain-specific QE model than that obtained in Step 2.

\subsection{Data augmentation for DA in QE}
\label{sec:DAG}
In our study, we explore two alternative approaches to oversampling to optimize the utilization of available ID resources and assess the potential benefits of incorporating synthetic ID data into the QE pipeline: 

\paragraph{Approach 1: Concatenating all available authentic ID data across all languages.}
The XLM-R model is multilingual, allowing us to apply it to different language pairs. When there is not enough data to fine-tune it for a specific language, one can use multilingual data. In our work, to increase the amount of authentic data (given the small volume of parallel data for two languages), we construct a multilingual ID dataset: we concatenate all available ID data, which includes different language pairs. 
The rationale behind this approach is to make use of all available authentic resources in order to improve the performance of the QE model by providing better cross-lingual references.

\paragraph{Approach 2: Generating synthetic ID data.}
Given that all available ID resources have been already utilized in Approach 1, we propose to supplement the existing data with artificially generated additional ID data using a trained MT model for each language pair, inspired by the research conducted by Negri et al.,~\shortcite{negri-etal-2018-escape} and Lee~\shortcite{lee-2020-two}. This approach aims to tackle the data scarcity problem and further improve the QE model's accuracy.
Let $D_{lp}$ denote the publicly available parallel data (SRC, TRG) for a language pair $lp$, as identified in Section~\ref{sec:data}. The approach consists of the following steps for each ID involved in the pipeline:
\begin{enumerate}[leftmargin=*,noitemsep]
    \item Randomly select $N$ samples from $D_{lp}$ to obtain a set $S_{lp}$ of training samples. Divide $S_{lp}$ into two equal sets $S_1$ and $S_2$.
    \item Train a multilingual MT model $M_{lp}$ on $S_1$ (details of the model can be found in Section~\ref{sec:MT}).
    \item Use $M_{lp}$ to translate the sources-side of $S_2$ (or a portion of it), obtaining a set $T_{lp}$ of translated samples.
    \item Compute quality labels (e.g., TER/HTER) by comparing $T_{lp}$ with the reference ($TRG$) text from $S_2$.
\end{enumerate}
The resulting three-part output of this approach comprises the source-side of $S_2$, $T_{lp}$, and TER/HTER obtained from the fourth step. A visual representation of these steps can be found in Appendix~\ref{appendix:approach2}.
 
\subsection{Additional indication of domain}
In NMT, in order to handle multiple domains and reduce catastrophic forgetting, DA has been controlled using additional tags added at the beginning or at the end of the sentence~\cite{sennrich-etal-2016-controlling,Chu2019MultilingualMA}.
Following these studies, we explore two training modes: (i) with tag (``TAG''), by appending either \verb|<OOD>| or \verb|<ID>| at the end of sentences based on the dataset domain type (i.e., OOD or ID). The input format in this mode is \verb|<s> SRC </s> TRG <Tag> </s>|, where SRC and TRG represent source and target of the QE triplet, and \verb|<s>| and \verb|</s>| are the beginning and separator tokens for the LLM used in the pipeline; (ii) without tag (``NO TAG''), where the training steps are the same as detailed in Section ~\ref{sec:training_steps}.



\vspace*{-1.0mm}
\section{Experiments}
\label{sec:experimental_setup}
\subsection{Data}
\label{sec:data}
We conducted experiments on publicly available data in different languages: from EN into DE, ZH, IT, CZ, and JA and from RO and RU into EN. We categorize the data into three groups according to their use in our pipeline: 
    

\paragraph{Group 1: for building \textit{ID} and \textit{OOD} QE models.} The \textit{ID} data is collected from WMT 2021 shared task on QE~\cite{specia-etal-2021-findings}, Task 2, consisting of sentence-level post-editing efforts for four language pairs: EN-DE, EN-ZH, RU-EN and RO-EN. For each pair there are train, development (dev), and test sets of 7$K$, 1$K$, 1$K$ samples, respectively. Additionally, as our \textit{OOD} data we used the eSCAPE~\cite{negri-etal-2018-escape} dataset with approximately 3.4$M$ tokenized SRC, machine-translated text~(MTT), post-edited~(PE) sentences. We used \texttt{sacrebleu}\footnote{signature:nrefs:1$|$case:lc$|$tok:tercom$|$punct:yes$|$version:2.3.1}~\cite{post-2018-call} to calculate TER~\cite{snover-etal-2006-study} from MTT and PE pairs. We split the data into train, dev, test sets via the \texttt{scikit-learn} package\footnote{random state/seed$=$8, shuffle$=$True, used for all splits.}~\cite{pedregosa2011scikit} with 98\%, 1\%, and 1\% of the total data, respectively.
To improve the generalization of our models and enable them to better adapt to specific QE through the ID dataset, we utilized a larger OOD dataset. This decision is in line with prior studies on DA, which are described in the related work section (Section~\ref{sec:related_works}).

\paragraph{Group 2: for building MT systems as a component of \textit{Approach 2} in the proposed DAG (Section~\ref{sec:DAG}).} We collected parallel data --- SRC and reference translations (REF) --- from Opus~\cite{tiedemann-2012-parallel} for each language pair used in ID: EN-DE, EN-ZH, RO-EN, and RU-EN. Next, we trained MT models for Approach 2 of our methodology by selecting 4$M$ samples and dividing them into two equal parts, each with 2$M$ samples. 
We split either of the two parts into train, dev, test sets. To save time during evaluation and inference, we set the size of the dev and test splits to be the same as the number of training samples in the ID datasets, which is 7$K$. 
Moreover, we randomly selected a portion of the SRC (7$K$ out of 2$M$) in the second split, which was not used for training. We passed this portion to the trained MT to get MTT. Finally, we computed the TER using the MTT and the corresponding REF via \texttt{sacrebleu}. We set the portion size 7$K$ as the goal was to double the size of the initial ID data. 

\paragraph{Group 3: for testing the zero-shot capabilities of the trained QE models in our proposed methodology.} We used two zero-shot test sets, namely English to Czech (EN-CS) and English to Japanese (EN-JA), which were provided by WMT 2021 shared task on QE for Task 2. Each test set contained 1$K$ samples. 

\subsection{Frameworks}
\paragraph{Quality Estimation.}
\label{sec:QE}
To train all QE models of our study, we developed a new QE framework with the ability to invoke multilingual models from HF model repository. In all our experiments we chose to use XLM-RoBERTa\footnote{xlm-roberta-large}~(XLM-R)~\cite{conneau-etal-2020-unsupervised}, to derive cross-lingual embeddings, which has shown success in prior studies such as~Ranasinghe et al.,~\shortcite{ranasinghe-etal-2020-transquest}. The framework is similar in architecture to ``MonoTransQuest''~\cite{ranasinghe-etal-2020-transquest}, but adapted to the needs of our experiments. The differences with ``MonoTransQuest'' are the additional tokens (\verb|<OOD>| and \verb|<ID>|) added  during the tokenization process
, as well as the resizing of the model's token embeddings in order to support the added tags. Additionally, rather than computing the softmax, we directly used logits to estimate the quality labels. 

\paragraph{Training and evaluation details of QE models.}
In Section~\ref{sec:training_steps} we describe our methodology for training and evaluating QE models. During Step 1, we trained and evaluated an OOD QE model every 1000 $steps_{HF}$\footnote{$steps_{HF}$ refers to Hugging Face framework's training or evaluation steps, which are different from the ones we described in Section~\ref{sec:training_steps}.} using the train and dev sets from Group 1. In Step 2, we trained and evaluated QE mix models every 500 $steps_{HF}$ using a mix of OOD and ID data from Group 1. For Step 3, we evaluated the final domain-specific QE model after 500 $steps_{HF}$ using only an ID train and dev set. Throughout training, we used an early stopping mechanism to halt the training process if there was no improvement in the evaluation loss after 5 evaluations. We adjusted the default evaluation $steps_{HF}$ from 500 to 1000 for Step 1 due to the larger number of training samples in that step.


\paragraph{Machine Translation.} 
\label{sec:MT}
Our approach to generating synthetic ID (Approach 2, Section~\ref{sec:DAG}) differs from prior studies, such as~Eo et al.,~\shortcite{https://doi.org/10.48550/arxiv.2111.00767}, which rely on a generic/common translation model (e.g., Google machine translate). Instead, we first trained a separate NMT model on a subset of the original dataset. This approach ensures that the training data and the data used for translation have similar vocabularies, cover comparable topics, styles, and domains, which leads to higher quality translations.


We used an in-house MT framework to train our models, based on pre-trained mBART-50 \cite{10.1162/tacl_a_00343} from HF. We followed the Seq2SeqTraining
arguments recommended by HF and trained the model for Approach 2, stopping the training if the evaluation loss did not improve after 5 evaluations.


We used default hyperparameters recommended by HF for QE and MT, and our frameworks with modified hyperparameters are available at~\url{https://github.com/JoyeBright/DA-QE-EAMT2023} to reproduce our results. 

\vspace*{-1mm}
\section{Results}
\label{sec:results}
To assess the performance of our approach we evaluate output from the trained QE models in comparison to the reference quality metric (HTER/TER) on the test sets described in data Groups 1 and 3.  We use Pearson's coefficient ($\rho \in -1:1$, which we rescale to $-100$ to $100$ for clarity) to correlate our predictions with the test set. We use the BLEU score as a metric to evaluate the translation quality of our MT models. 

\subsection{Baseline results}
\label{sec:baseline}
To establish a baseline for our study, we fine-tuned XLM-R with the ID data for each language pair as provided by WMT 2021 shared task (Group 1 of data). This is a conventional approach employed in prior research, such as~Ranasinghe et al.~\shortcite{ranasinghe-etal-2020-transquest}, where pre-trained models are utilized to provide cross-lingual reference for training QE models. 

We also attempted to compare our work with the models of Rubino~\shortcite{rubino-2020-nict} and Lee~\shortcite{lee-2020-two}. For the latter work, their experiments used the WMT 2020 test sets, while we used WMT 2021, which makes it difficult to compare our results to theirs directly. Furthermore, we could not replicate their models as no code is available (at the time of writing this paper). Our baseline results are presented in Table~\ref{table:main_results}.

\subsection{Main results}
In Table~\ref{table:main_results} we present our results using the DAG approaches and the two training modes (Tag and No Tag). Additional details on the statistical tests for each language pair are available in Appendix~\ref{appendix:ss}. The results in Table~\ref{table:main_results} show that, in general, all of the proposed DA methods performed better than the baseline for each language pair, except for Approach 1 in the RO-EN language pair. For this language pair, the use of a domain tag led to reduced performance, and the improvement achieved without such a tag was not statistically significant.

\begin{table}
\setlength\tabcolsep{3pt}
\renewcommand{\arraystretch}{1.2}
\centering
\begin{adjustbox}
{width=\columnwidth,center}
\begin{tabular}{l|c|cc|cc|cl}
\hlineB{3}
\multirow{2}{*}{\pbox{20cm}{Language \\ pair}} &
  \multirow{2}{*}{Baseline} &
  \multicolumn{2}{c}{NO TAG} &
  \multicolumn{2}{c|}{TAG} &
  \multicolumn{2}{c}{\multirow{2}{*}{Increase \%}} \\ \cline{3-6}
 &
   &
  \multicolumn{1}{l}{DAG 1} &
  \multicolumn{1}{l|}{DAG 2} &
  \multicolumn{1}{l}{DAG 1} &
  \multicolumn{1}{l|}{DAG 2} &
  \multicolumn{2}{c}{} \\ \hline
EN-DE        & 47.17 & \multicolumn{1}{c}{49.93} & 49.54 & \multicolumn{1}{c}{\textbf{51.90}}  & 51.25 & \multicolumn{2}{c}{10.03} \\ 
EN-ZH        & 29.16 & \multicolumn{1}{c}{34.75} & 35.27 & \multicolumn{1}{c}{35.62} & \textbf{36.60}  & \multicolumn{2}{c}{25.51} \\ 
RO-EN        & 83.63 & \multicolumn{1}{c}{83.67} & 83.74 & \multicolumn{1}{c}{83.37} & \textbf{84.40}  & \multicolumn{2}{c}{00.92}  \\ 
RU-EN        & 40.65 & \multicolumn{1}{c}{44.91} & 45.40 & \multicolumn{1}{c}{\textbf{47.16}} & 43.98 & \multicolumn{2}{c}{16.01}  \\ \hlineB{3}
\end{tabular}
\end{adjustbox}
\caption{\textbf{Pearson correlation scores for proposed QE models across 4 language pairs}: EN-DE, EN-ZH, RO-EN, and RU-EN. For each language pair, the bold result indicates the highest-performing method compared to the baseline. Results for the first and second DAG approaches are reported under DAG 1 and DAG 2, respectively. The column labeled ``Increase \%" shows the percentage improvement for the highest-performing model (in bold) compared to the baseline.}

\label{table:main_results}
\end{table}

We also observe that the increase of performance compared to the baseline for each language pair shown as percentage in the last column of Table~\ref{table:main_results} is substantial, except for RO-EN (only 0.92\% increase over the baseline). This is mainly due to the already high baseline performance (83.63), making it challenging to achieve significant improvements. Among the other language pairs, the EN-ZH pair had the largest increase in performance –– just over 25\%. The RU-EN and EN-DE pairs had the second and third highest increases, with improvements of around 16\% and 10\% over their respective baselines. 

\paragraph{Additional indication of domain results.}
The results indicate that incorporating tags into the DA training pipeline was generally effective, although in some instances, the improvement was not statistically significant compared to the models that were trained without tags. However, it was observed that at least one model outperformed the same language pair's models that were not trained with tags, when DAG techniques were used. Specifically, the EN-DE Approach 1 model trained with tags performed better compared to Approach 2 without tags, as did the EN-ZH Approach 1 model trained with tags relative to the same approach without tags. Finally, the RO-EN Approach 2 model trained with tags outperformed Approach 2 without tags, and the RU-EN Approach 1 model trained with tags exhibited better performance than Approach 1 without tags. 

\vspace*{-1mm}
\subsection{Data Augmentation results}
Upon analyzing the integration of DAG techniques into the specialized QE pipeline, we observe that for most language pairs, both approaches showed better performance than their respective baselines. However, in situations where tags were not employed, Approach 2 only showed statistical significance over Approach 1 in the EN-ZH and RU-EN language pairs. Moreover, when tags were used, Approach 2 lead to statistically significant improvements only for EN-DE and EN-ZH. These findings suggest that the choice of DAG approach and the use of tags should be carefully considered when applying DA in QE. Additionally, DAG was observed to be significant for EN-ZH, for both cases --- with or without tags. 


\vspace*{-2.00mm}
\subsection{Zero-shot results}
In order to evaluate the effectiveness of our QE models in the context of ZSL, we compared their performance with the baseline models for the EN-CS and EN-JA language pairs (test sets). The results of these tests are presented in Table~\ref{tbl:zsl}.

The findings show that, for the EN-CS test set, the QE model trained solely on the EN-DE dataset achieved the highest performance among all QE baselines, with a Pearson correlation score of 46.97. Additionally, we observe that our proposed DA pipeline performed even better than the highest-performing baseline for EN-CS, but only DAG approach 1 and 2 with tags were found to be statistically significant. Likewise, for the EN-JA test set, the highest-performing QE baseline was the one that was trained solely on the RU-EN dataset, with a Pearson correlation score of 20.32. In contrast to EN-CS, none of the models that were trained with our pipeline and with the RU-EN dataset outperformed the baselines. Nevertheless, we observed that three models trained with EN-ZH and using our pipeline (Approach 1 with and without tag, and Approach 2 with tag) performed better than the highest-performing baseline.

Overall, these findings suggest that if a QE model is conventionally trained with and evaluated on an unseen QE dataset, some extent of ZSL capabilities can be achieved due to the use of XLM-R. However, the proposed DA pipeline can significantly increase this extent, whether through models trained with the same dataset or other datasets used in the pipeline. Furthermore, we observed that training a QE model conventionally using certain language pairs may lead to decreased performance. For instance, a model trained exclusively with the EN-DE language pair showed a Pearson correlation of approximately 10. In such cases, the proposed pipeline may enhance performance even when using the same training data. 

\begin{table}[h]
\setlength\tabcolsep{5pt}
\renewcommand{\arraystretch}{1.1}
\centering
\begin{adjustbox}
{width=\columnwidth,center}
\begin{tabular}{l|c|c|cc|cc}
\hlineB{3}
\multirow{2}{*}{\begin{tabular}[c]{@{}l@{}}Trained \\ on\end{tabular}} &
  \multirow{2}{*}{Test set} &
  \multirow{2}{*}{Baseline} &
  \multicolumn{2}{c|}{NO   TAG} &
  \multicolumn{2}{c}{TAG} \\ \cline{4-7} 
                       &       &       & DAG 1 & DAG 2 & DAG 1 & DAG 2 \\ \hline\hline
\multirow{2}{*}{EN-DE} & EN-CS & 46.97 & 48.77 & 48.07 & 47.78 & 47.82 \\
                       & EN-JA & 09.67  & 18.16 & 08.00 & 16.12 & 17.36 \\ \hline\hline
\multirow{2}{*}{EN-ZH} & EN-CS & 35.56 & 49.33 & 48.54 & 47.98 & 46.83 \\
                       & EN-JA & 13.13 & 22.77 & 19.87 & 22.24 & 21.54 \\ \hline\hline
\multirow{2}{*}{RO-EN} & EN-CS & 26.33 & 39.10 & 39.79 & 39.20 & 40.41 \\
                       & EN-JA & 18.88 & 20.34 & 18.55 & 20.11 & 21.22 \\ \hline\hline
\multirow{2}{*}{RU-EN} & EN-CS & 28.42 & 45.58 & 44.85 & 46.43 & 45.22 \\
                       & EN-JA & 20.32 & 17.64 & 17.04 & 17.26 & 19.63 \\ \hlineB{3}
\end{tabular}
\end{adjustbox}
\caption{\textbf{Performance comparison of the proposed methods and the baseline model} trained on the EN-DE, EN-ZH, RO-EN, and RU-EN datasets in the context of ZSL, with results presented for EN-CS and EN-JA test sets. Results for the first and second DAG approaches are reported under DAG 1 and DAG 2, respectively.}
\label{tbl:zsl}
\end{table}
\vspace*{-1.00mm}
\section{Additional observations}
\label{sec:discussion}
\subsection{Cross-lingual inference}

Table~\ref{table:cross-lingual} presents data that shows that our proposed methodology has an overall advantage over the conventional training method of using a pre-trained LLM and fine-tuning it with QE data (baselines) in terms of cross-lingual inference. That is, the QE models trained with our proposed DA pipeline not only perform significantly better than baselines on their target domain and language pair but can also estimate the quality of other language pairs to some extent better than their corresponding baseline.

By examining the data closely (bottom to top row of the Table~\ref{table:cross-lingual}), we observe that XLM-R provides a limited level of cross-lingual inference, which is insufficient for estimating quality labels due to the absence of prior knowledge about them. However, using Step 1 of our pipeline, which utilizes little inference knowledge, the model still achieves an acceptable level of generalization across all language pairs. 

Specifically, the first step achieved an average Pearson correlation score of approximately 39, which is higher than all baseline scores, except for the RO-EN pair, which achieved around 42. Furthermore, the model trained using Step 1 of the pipeline achieved a Pearson correlation of around 70 when evaluated with the RO-EN test set. This result can be attributed to the training of the model with IT, which was used as OOD data. From a linguistic point of view, this result could be explained by the fact that IT and RO belong to the same language family, i.e., the ``romance languages'' (refer to Appendix~\ref{appendix:genetic}), which explains the high Pearson correlation score achieved by the model.

As we move up the table, we can observe that the model built in Step 2 of our pipeline becomes more specific toward the task and the ID datasets. Consequently, there is an average improvement of around 3.5 Pearson correlation (from 39.36 to 42.83) across the languages. This indicates that our DA pipeline is effective in improving more specific cross-lingual QE performance. Ultimately, fine-tuning Step 2 with any of the ID languages provides a highly domain-specific QE model that is not only better estimates the quality of their language pair, but also performs better cross-lingual inference over its baseline.

\begin{table}[t]
\renewcommand{\arraystretch}{1.05}
\centering
\begin{adjustbox}
{width=\columnwidth,center}
\begin{tabular}{l|cccc|c}
\hlineB{3}
  \multirow{2}{*}{Models} &
  \multicolumn{4}{c|}{Test Sets} &
  \multirow{2}{*}{AVG}
  \\ \cline{2-5}
                                          & EN-DE & EN-ZH & RO-EN & RU-EN & \\ \hline\hline
\cellcolor[HTML]{FFFF00}Baseline & \underline{47.17}          & 19.67          & 44.96          & 32.91          & 36.17             \\
\cellcolor[HTML]{FFFF00}EN-DE    & \underline{49.93}          & 22.66          & 78.97          & 39.55          & 47.77             \\
\textbf{$\Delta$}                         & \underline{02.76}           & 02.99           & 34.01          & 06.64           & \textbf{11.60}       \\ \hline\hline
\cellcolor[HTML]{FFE699}Baseline & 30.34          & \underline{29.16}          & 47.55          & 36.87          & 35.98            \\
\cellcolor[HTML]{FFE699}EN-ZH    & 43.46          & \underline{34.75}          & 80.51          & 42.67          & 50.34             \\
\textbf{$\Delta$}                         & 13.12          & \underline{05.59}           & 32.96          & 05.80            & \textbf{14.36}      \\ \hline\hline
\cellcolor[HTML]{A9D08E}Baseline & 24.64          & 23.56          & \underline{83.63}          & 39.97          & 42.95              \\
\cellcolor[HTML]{A9D08E}RO-EN    & 43.02          & 24.31          & \underline{83.67}          & 38.74          & 47.43             \\
\textbf{$\Delta$}                         & 18.38          & 00.75           & \underline{00.04}           & -01.23          & \textbf{04.48}        \\ \hline\hline
\cellcolor[HTML]{F4B084}Baseline & 22.40           & 24.67          & 57.17          & \underline{40.69}          & 36.23              \\
\cellcolor[HTML]{F4B084}RU-EN    & 25.36          & 26.06          & 75.34          & \underline{44.91}          & 42.91             \\
\textbf{$\Delta$}                         & 02.96           & 01.39           & 18.17          & \underline{04.22}           & \textbf{06.68}        \\ \hline\hline
Step2                            & 38.29          & 24.72          & 76.96          & 31.35          & 42.83 
\\  \hline
Step1                            & 30.80           & 16.57          & 70.14          & 39.93          & 39.36     
 \\  \hline
XLM-R  &-02.74	&07.30	&02.97	&03.12	&02.66  
\\
\hlineB{3}
\end{tabular}
\end{adjustbox}
\caption{\textbf{Performance comparison of proposed models and baselines across all test sets} using Pearson correlation as the metric. $\Delta$ represents the difference between them. ``AVG'' column shows the overall difference for each language model. Step 1: model trained with OOD. Step 2: model trained with DAG approach 1 and OOD. Approach 2 in Step 2 had similar results, not included. XLM-R: model not being trained. Models and baselines are color-coded for clarity, with bold numbers indicating the average $\Delta$ across all language pairs, and underlined numbers representing each model's performance on their respective test sets.}
\label{table:cross-lingual}
\end{table}
\vspace*{-1mm}
\subsection{OOD Performance}
The main goals of DA are to quickly create an adapted system and to develop a system that performs well on ID test data while minimizing performance degradation on a general domain. In our study, we showed that models from Step 1 or Step 2 can be fine-tuned quickly using the user's data (achieving the first of these goals). Our main focus was on the assessment of ID QE. However, we test the generalizability of our ID models  on an OOD test set. Our results, summarized in Table~\ref{table:OOD_performance}, indicate that all ID models outperformed the corresponding baselines on the OOD test set, and we observe that incorporating ID data in Approaches 1 and 2 did not compromise the performance with respect to OOD. However, comparing the models' performance with models trained solely on OOD we see a small performance drop, which is inevitable and in most cases acceptable.
\begin{table}[ht]
\setlength\tabcolsep{2pt}
\renewcommand{\arraystretch}{1.15}
\centering
\begin{adjustbox}
{width=\columnwidth,center}
\begin{tabular}{l|cccc|c|c|c}
\hlineB{3}
\multirow{2}{*}{\pbox{20cm}{Trained \\ with}} &
  \multicolumn{7}{c}{QE Models}
  \\ \cline{2-8}
          & EN-DE & EN-ZH & RO-EN & RU-EN & OOD  & DAG 1    & DAG 2    \\ \hline
Baseline & 11.95          & 03.59           & 11.60           & 03.43           & \multirow{4}{*}{64.33} & \multirow{4}{*}{65.24} & \multirow{4}{*}{64.76} \\ 
Our pipeline           & 54.62 & 59.30  & 52.51  & 47.36  &  &  &  \\ \cline{1-5}
$\Delta_{Baseline}$ & 42.67 & 55.71 & 40.91  & 43.93  &  &  &  \\
$\Delta_{OOD}$  & \textbf{-09.71} & \textbf{-05.03} & \textbf{-11.82} & \textbf{-16.97} &  &  & \\ \hlineB{3}
\end{tabular}
\end{adjustbox}
\caption{\textbf{Model comparison on OOD test set} using Pearson correlation as the metric. The $\Delta_{Baseline}$ values indicate the performance difference relative to the corresponding baseline, while the $\Delta_{OOD}$ values compare the models' performance with the one trained solely with OOD.}
\label{table:OOD_performance}
\end{table}

\vspace*{-2.50mm}
\section{Related Work}
\label{sec:related_works}
\vspace*{-1.0mm}


\paragraph{Data Scarcity in QE.} The issue of data scarcity in MT QE has been explored in numerous previous studies. The work of Rubino and Sumita~\shortcite{rubino-sumita-2020-intermediate} involves the use of pre-training sentence encoders and an intermediate self-supervised learning step to enhance QE performances at both the sentence and word levels. This approach aims to facilitate a smooth transition between pre-training and fine-tuning for the QE task. 
Similarly, Fomicheva et al.,~\shortcite{fomicheva-etal-2020-unsupervised} proposed an unsupervised method for QE that does not depend on additional resources and obtains valuable data from MT systems. 

Qiu et al.~\shortcite{2022arXiv221210257Q} conducted a recent study on the the impact of various types of parallel data in QE DAG, and put forward a classifier to differentiate the parallel corpus. Their research revealed a significant discrepancy between the parallel data and real QE data, as the most common QE DAG technique involves using the target size of parallel data as the reference translation~\cite{baek-etal-2020-patquest,2022arXiv221210257Q}, followed by translation of the source side using an MT model, and ultimately generating pseudo QE labels~\cite{freitag-etal-2021-results}. However, our study diverges from this conventional approach and concentrates on a straightforward yet effective DAG methods to mitigate this gap. Similarly, Kocyigit et al.~\shortcite{kocyigit-etal-2022-better} proposed a negative DAG technique to improve the robustness of their QE models. They suggested training a sentence embedding model to decrease the search space and training it on QE data using a contrastive loss.
\vspace*{-1mm}
\paragraph{Domain Adaptation in QE.} To tackle the challenges with translating data when training data comes from diverse domains, researchers have extensively used DA in MT. DA involves training a large generic model and then fine-tuning its parameters with domain-specific data~\cite{chu-wang-2018-survey,Saunders2021DomainAA,2021arXiv211206096P,pham-etal-2022-multi}. In MT, one way to achieve DA is by appending tags to sentences to handle different domains~\cite{sennrich-etal-2016-controlling,vanmassenhove-etal-2018-getting,Chu2019MultilingualMA} and reduce catastrophic forgetting. 

Despite being useful in MT, DA has not been widely used in QE according to our knowledge. Dongjun Lee~\shortcite{lee-2020-two} proposed a two-step QE training process similar to our own, and Raphael Rubino~\shortcite{rubino-2020-nict} pre-trained XLM and further adapted it to the target domain through intermediate training. Both studies demonstrated that adding a step before fine-tuning improves performance compared to fine-tuning alone. However, unlike our methodology, neither of them included sentence tags or conducted additional fine-tuning (such as Step 3 in our methodology). As a result, their QE models are not as specialized for the target domain as ours.
A few researchers have made attempts to integrate aspects of DA into QE. For instance, in an effort to improve QE performance in domain-specific scenarios, Arda Tezcan~\shortcite{Tezcan_2022} included fuzzy matches into MonoTransQuest with the aid of XLM-RoBERTa model and data augmentation techniques. 

\vspace*{-2mm}
\section{Conclusion and future work}
\label{sec:conclusion}
This paper addresses two key challenges related to quality estimation~(QE) of machine translation~(MT): (i) the scarcity of available QE data and (ii) the difficulties in estimating translations across diverse domains. The primary aim of this study is to enhance the performance of QE models by addressing these challenges. To do so, we propose a solution that utilizes domain adaptation~(DA) techniques adopted from MT. We adapt the ``mixed fine-tuning + fine-tuning'' approach~\cite{chu-etal-2017-empirical} and extend it with data augmentation as an alternative to the traditional oversampling technique. We adopt a three-step training methodology: (i) we fine-tune XLM-R, a language model, with a large generic QE dataset, which enables the model to generalize; (ii) we fine-tune the model with a mix of out-of-domain~(OOD) and in-domain~(ID) data derived from two data augmentation~(DAG) approaches; and (iii) we fine-tune the model with a small amount of domain-specific data, which leads to a more specific model. We evaluated models' performance with and without domain tags appended to the sentences.

Our experiments show significant improvements across all language pairs under consideration, indicating that our proposed solution has a beneficial impact in addressing the aforementioned challenges. Our study also demonstrates the effectiveness of both proposed DAG approaches and shows that using domain tags improves the performance of the models. Additionally, we find that our model outperforms the baseline in the context of zero-shot learning and in cross-lingual inference. 

Moving forward, there are several directions for future work based on our findings. First, it would be interesting to investigate the performance of our pipeline on low-resource language pairs, where there is limited ID data available. This is particularly relevant given the smaller coverage of QE datasets compared to parallel data in MT. Second, we only used one type of OOD data in our experiments (EN-IT); it would be useful to explore other OOD data over different language pairs for QE. Third, it would be valuable to study the performance of other LLMs than XLM-R. Fourth, since the choice of languages employed in the pipeline was based on availability, we would suggest exploring a more regulated approach for selecting the languages to be used in the proposed pipeline. Specifically, the optimal transfer languages can be selected based on their data-specific features, such as dataset size, word overlap, and subword overlap, or dataset-independent factors, such as genetic (see Appendix~\ref{appendix:genetic}) and syntactic distance~\cite{lin-etal-2019-choosing}. 




\vspace*{-5mm}
\bibliography{eamt23}
\bibliographystyle{eamt23}

\appendix
\section{Appendices}

\subsection{Training Steps}
\begin{figure}[h]
    \centering
     \includegraphics[keepaspectratio,width=\columnwidth]{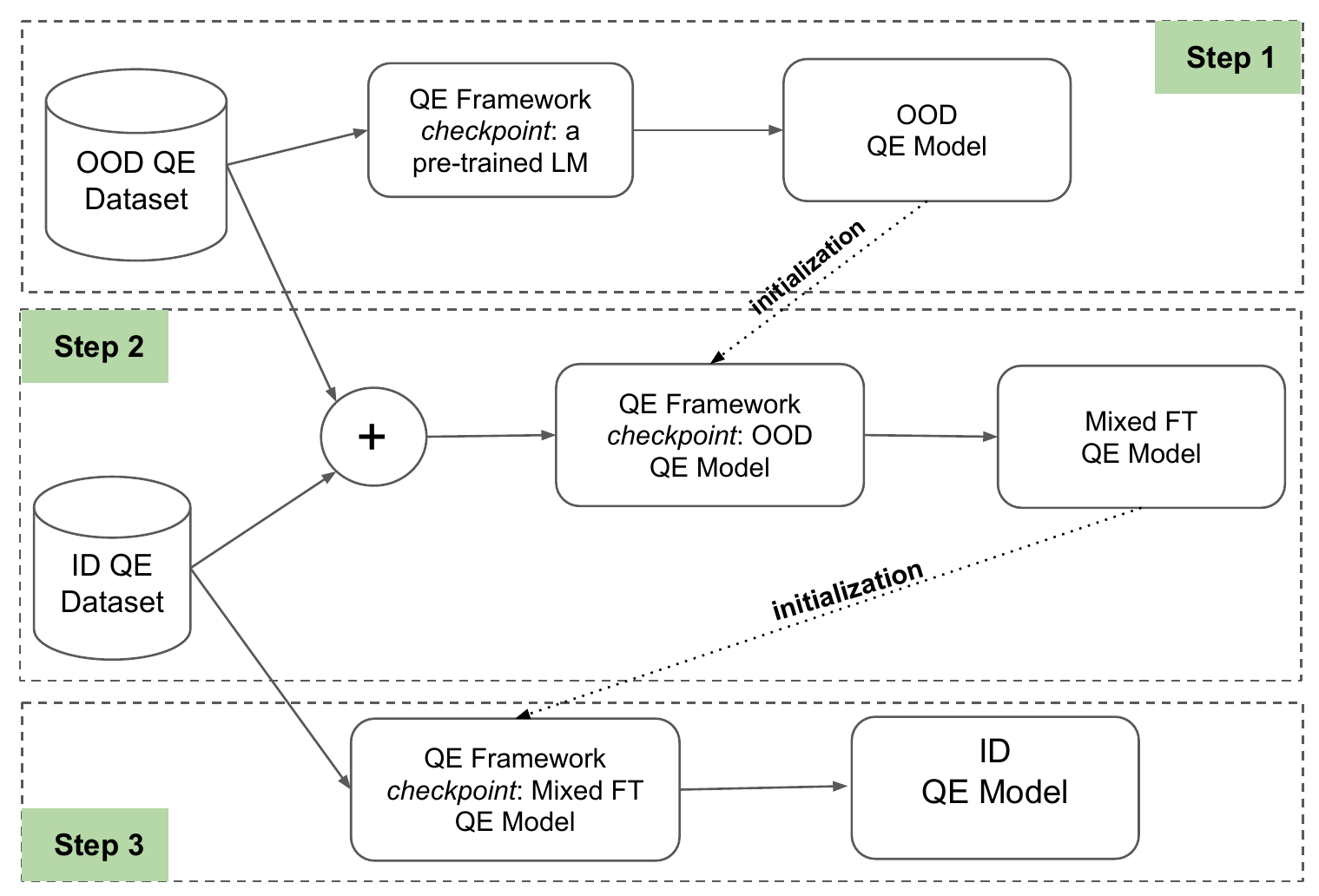}
    \caption{\textbf{Overview of the proposed training steps for specialized QE.} The ``+'' sign indicates the oversampling performed in Step 2 to balance the use of ID and OOD data. The dashed arrows indicate the source of the checkpoint used to initialize the models in each stage.}
    \label{fig:training_Pipeline}
\end{figure}

In Figure~\ref{fig:training_Pipeline}, we present an overview of the proposed training steps for specialized QE.
\label{appendix:training_steps}

\subsection{Statistically Significance Test Results}
\label{appendix:ss}

The statistical significance test results for the predictions in Table~\ref{table:main_results} for the language pairs EN-DE, EN-ZH, RO-EN, and RU-EN are shown in Table~\ref{tab:ss_all}.

\begin{table}[h]
\setlength\tabcolsep{3pt}
\renewcommand{\arraystretch}{1.0}
\centering
\begin{adjustbox}
{width=\columnwidth,center}
\begin{tabular}{l|l|cccc}
\hlineB{3}
\begin{tabular}[c]{@{}l@{}}Language   \\ pair\end{tabular} & \multicolumn{1}{l|}{Models} & NO TAG 1 & NO TAG 2 & TAG 1 & TAG 2 \\ \hline\hline
\multirow{4}{*}{\textbf{EN-DE}} & Baseline & Y & Y & Y & Y \\
                                & NO TAG 1 & - & N & N & Y \\
                                & NO TAG 2 & - & - & Y & Y \\
                                & TAG 1    & - & - & - & Y \\ \hline\hline
\multirow{4}{*}{\textbf{EN-ZH}} & Baseline & Y & Y & Y & Y \\
                                & NO TAG 1 & - & Y & Y & N \\
                                & NO TAG 2 & - & - & N & N \\
                                & TAG 1    & - & - & - & Y \\ \hline\hline
\multirow{4}{*}{\textbf{RO-EN}} & Baseline & N & Y & Y & Y \\
                                & NO TAG 1 & - & N & Y & Y \\
                                & NO TAG 2 & - & - & N & N \\
                                & TAG 1    & - & - & - & N \\ \hline\hline
\multirow{4}{*}{\textbf{RU-EN}} & Baseline & Y & Y & Y & Y \\
                                & NO TAG 1 & - & Y & Y & Y \\
                                & NO TAG 2 & - & - & N & Y \\
                                & TAG 1    & - & - & - & N \\ \hlineB{3}
\end{tabular}
\end{adjustbox}
\caption{\textbf{Statistically significant test results} with a p-value less than 0.05. The letter ``Y" in the table indicates that the corresponding prediction in Table~\ref{table:main_results} is statistically significant, while ``N" indicates that it is not.}
\label{tab:ss_all}
\end{table}

\subsection{Data Augmentation: Approach 2}
\begin{figure}[htbp]
    \centering
    \includegraphics[keepaspectratio,width=\columnwidth]{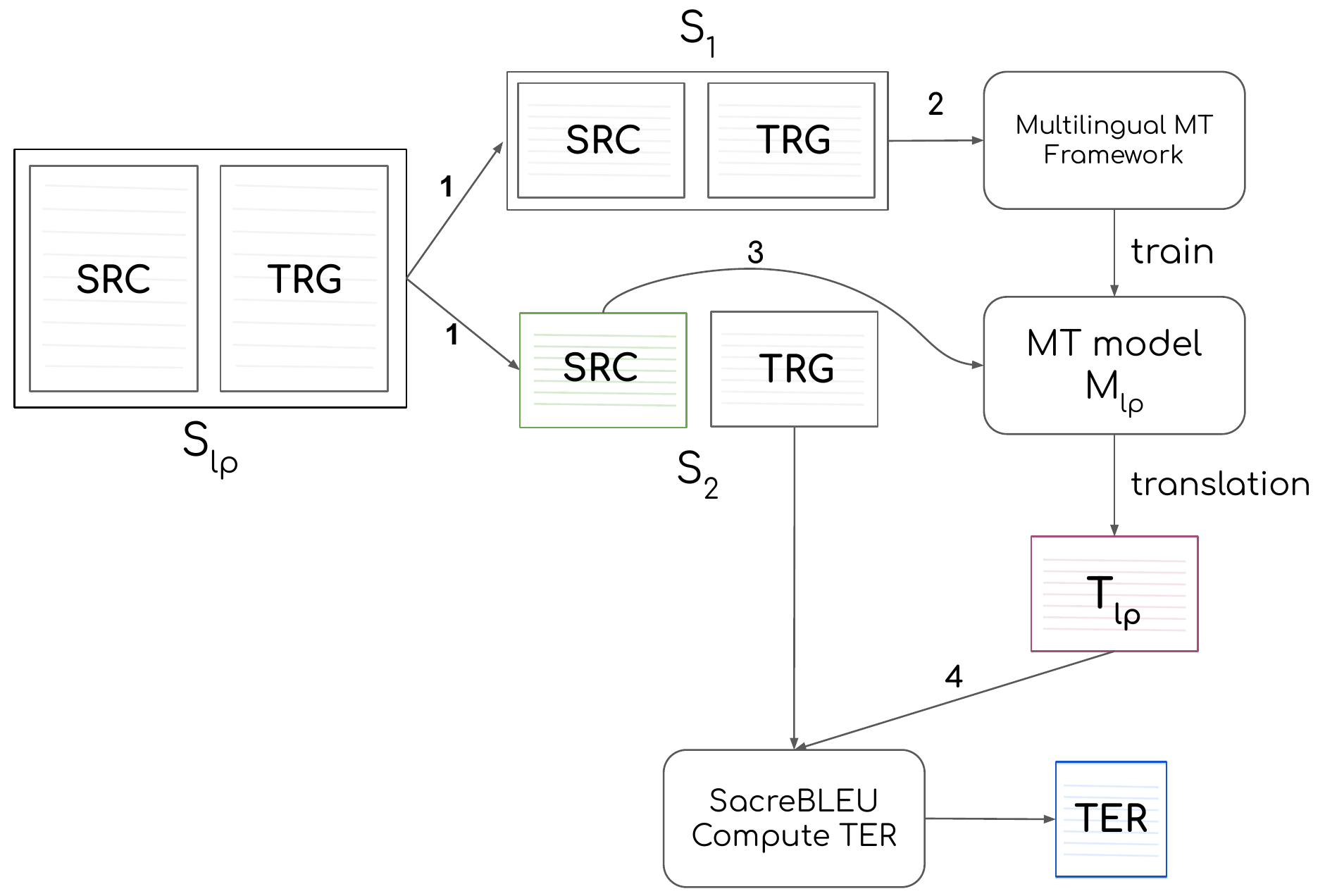}
    \caption{\textbf{Overview of Approach2 (Generating synthetic ID) of data augmentation for domain adaptation in QE.} The various steps involved in the approach are indicated close to the corresponding arrows. Arrow 1 represents subsampling. The abbreviations $SRC$, $TRG$, and $T_{lp}$ stand for source, target, and machine-translated text, respectively. The final outputs which include $SRC$, $T_{lp}$ and quality labels ($TER$) are color-coded for clarity.}
    \label{fig:approach2}
\end{figure}

Figure~\ref{fig:approach2} presents an overview of Approach 2 that is employed for data augmentation in the context of domain adaptation for QE.
\label{appendix:approach2}
\subsection{Machine Translation Performance}
We utilized multilingual MT systems to generate synthetic ID data. Table~\ref{tab:mt_DAG2} displays the results of the top-performing models used in generating this data.
\begin{table}[h]
\setlength\tabcolsep{6pt}
\renewcommand{\arraystretch}{1.0}
\centering
\begin{adjustbox}
{width=160pt,center}
    \begin{tabular}{l|c|c}
        \hlineB{3}
        Language pair & BLEU~$\uparrow$ & Eval Loss~$\downarrow$ \\ \hline
        EN-DE & 41.25 & 01.09 \\
        EN-ZH & 32.28 & 01.52 \\
        RO-EN & 49.60 & 00.96 \\
        RU-EN & 41.29 & 01.61 \\  \hlineB{3}
    \end{tabular}
\end{adjustbox}
    \caption{\textbf{MT performance used as a component of Approach 2} in the proposed DAG~(Section~\ref{sec:DAG}).}
    \label{tab:mt_DAG2}
\end{table}

\subsection{Genetic Distance}
\label{appendix:genetic}
\begin{figure}[h]
    \centering
    \includegraphics[width=130px]{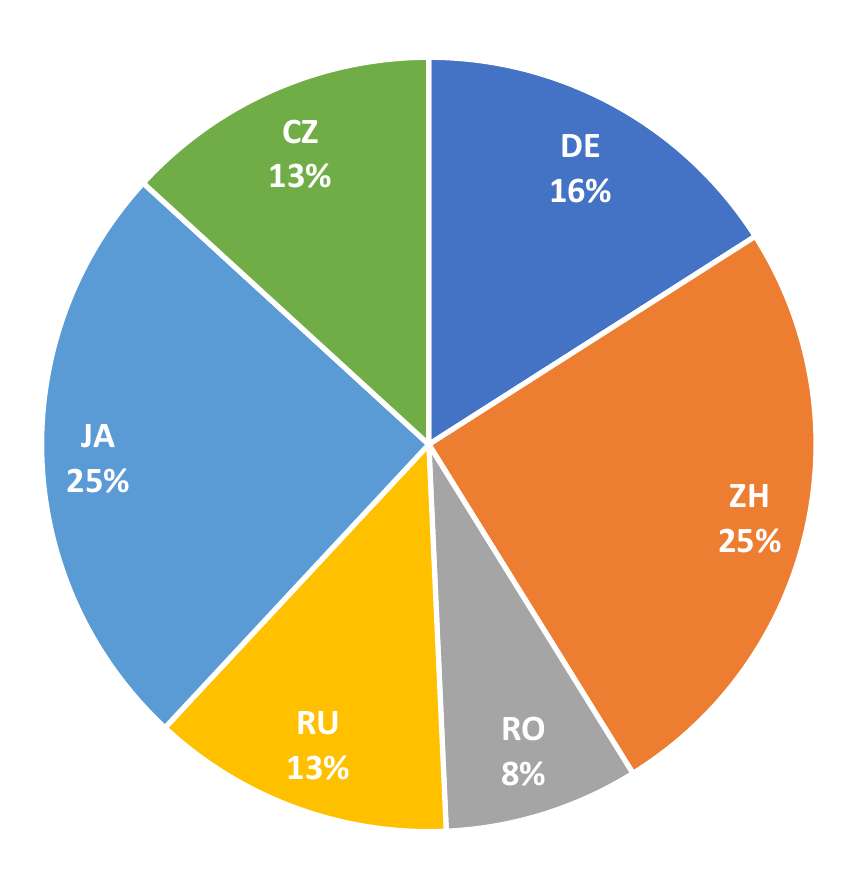}
    \caption{\textbf{Genetic distance between IT and other languages:} DE, ZH, RO, RU, JA, and CZ.}
    \label{fig:genetic}
\end{figure}
In MT, measuring the similarity between languages is important for effective cross-lingual learning. One such measure is the ``genetic distance'' between languages, which has been shown to be a good indicator of language similarity for independent data~\cite{lin-etal-2019-choosing}. To illustrate this, we calculate\footnote{\url{http://www.elinguistics.net/Compare_Languages.aspx}} and present the genetic distance scores between Italian (used as OOD data) and the other languages included in our study in Figure~\ref{fig:genetic}. The genetic distance is represented as a numerical value ranging from 0 (indicating the same language) to 100 (the greatest possible distance).

\subsection{Training time}
\begin{figure}[t]
    \centering
    \includegraphics[width=1.0\columnwidth]{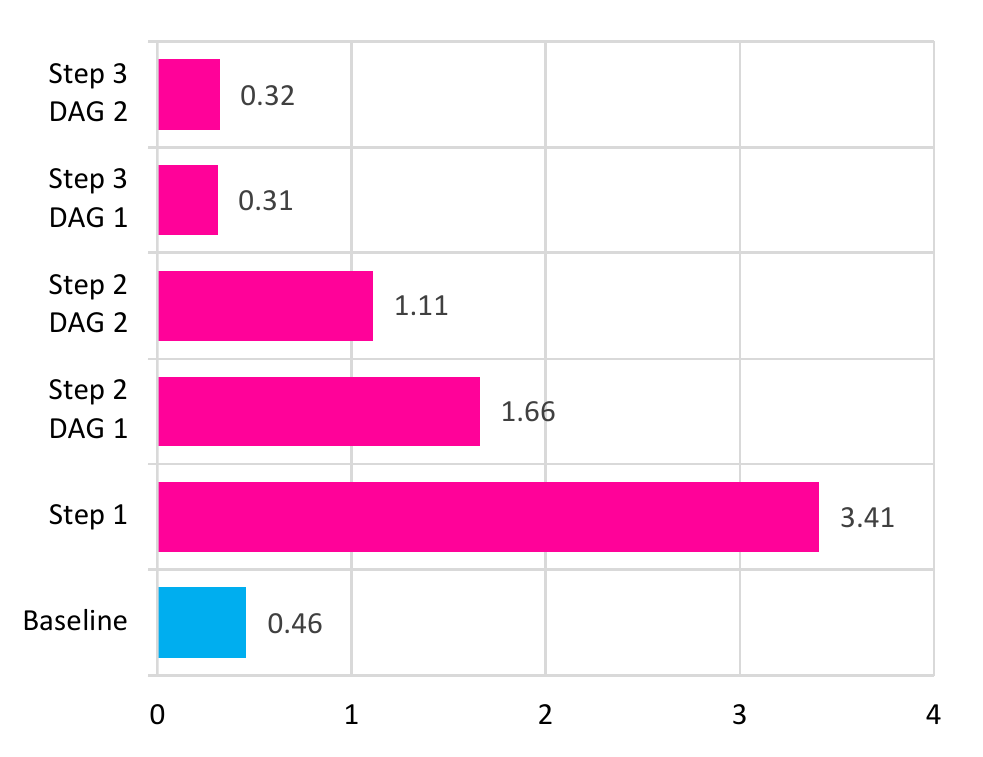}
    \caption{\textbf{Training time (in hours) for models in the EN-ZH language pair}, where Step X refers to the training step outlined in Section~\ref{sec:training_steps}, and DAG X denotes the data augmentation approach used in the second step of the pipeline. The term ``Baseline'' denotes a model fine-tuned from XLM-R. The X and Y axes represent the training time in hours and the approaches used to train the model, respectively.}
    \label{fig:training_time}
\end{figure}

Compared to the conventional approach of using a pre-trained LLM and fine-tuning it with QE data (baselines), our proposed DA methodology results in a significant improvement in performance, regardless of whether we include tags in the sentences or not. However, it requires two additional training steps: Step 1, training an OOD QE model, and Step 2, fine-tuning the model using a mix of OOD and ID QE data. These additional steps require more time. Step 1 and Step 2 (with both DAG approaches) are reused (i.e., not trained) for each language pair, and Step 3 of the pipeline took almost the same amount of time across all languages. That is why we present the consumed time for EN-ZH in Figure~\ref{fig:training_time}, and use it to discuss training times for other language pairs as well. Models trained with tagged data have a similar training time.

The data presented in Figure~\ref{fig:training_time} indicates that Step 1 has the highest training time with approximately 3.4 hours. It is noteworthy that this long training time is partly due to the fact that the model was evaluated after every 1000 $steps_{HF}$, which consequently resulted in a longer running time in comparison to other models that were evaluated after every 500 $steps_{HF}$. Furthermore, the model that was trained is publicly accessible, and other individuals can utilize it to fine-tune with new ID datasets, avoiding the need for retraining for each specific ID data. This applies to both DAG approaches, given that the target language pair was used in Step 2 of the pipeline. If not, Step 1 must be fine-tuned with a new set of QE data.

\end{document}